\documentclass[runningheads]{llncs}
\usepackage{graphicx}
\usepackage{subcaption}
\usepackage[dvipsnames]{xcolor}
\usepackage{array}
\newcolumntype{P}[1]{>{\centering\arraybackslash}p{#1}}
\begin{document}
\title{A Transfer Learning Based Model for Text Readability Assessment in German}
\titlerunning{A Transfer Learning Model for German Text Readability}
% \author{Anonymous Authors%\inst{1}%\orcidID{0000-1111-2222-3333} %\and
%Second Author\inst{2,3}\orcidID{1111-2222-3333-4444} \and
%Third Author\inst{3}\orcidID{2222--3333-4444-5555}
\author{
Salar Mohtaj\inst{1,2} \and %\orcidID{0000-0002-0032-3833} \and 
Babak Naderi\inst{1} \and
Sebastian Möller\inst{1,2} \and \\
Faraz Maschhur\inst{1} \and
Chuyang Wu\inst{1} \and
Max Reinhard\inst{1}
}%
\institute{
Technische Universit\"at Berlin, Berlin, Germany\\
\and
German Research Centre for Artificial Intelligence (DFKI), Labor Berlin, Germany \\
\email{\{salar.mohtaj $\mid$ babak.naderi $\mid$ sebastian.moeller\}@tu-berlin.de}\\ 
}
\authorrunning{Salar Mohtaj et al.}

%\institute{Princeton University, Princeton NJ 08544, USA \and
%Springer Heidelberg, Tiergartenstr. 17, 69121 Heidelberg, Germany

%
\maketitle              % typeset the header of the contribution
\begin{abstract}
Text readability assessment has a wide range of applications for different target people, from language learners to people with disabilities. The fast pace of textual content production on the web makes it impossible to measure text complexity without the benefit of machine learning and natural language processing techniques. Although various research addressed the readability assessment of English text in recent years, there is still room for improvement of the models for other languages. In this paper, we proposed a new model for text complexity assessment for German text based on transfer learning. Our results show that the model outperforms more classical solutions based on linguistic features extraction from input text. The best model is based on the BERT pre-trained language model achieved the Root Mean Square Error (RMSE) of 0.483.

\end{abstract}
\section{Introduction}
\label{sec:introduction}
Text forms an integral part of exchanging information and interacting with the world. Social media and web accelerated the textual content production compared to the era before the web. In other words, along with the other types of content (e.g., image and video), textual content has been increasing drastically during past recent years. Text readability (in the following used interchangeably with text complexity) is one of the factors which affects a reader's understanding of text~\cite{dale1949concept}. \par
A readability score is significant in informing readers about the difficulty of a piece of text (e.g., a document) that they read. A readability score is the mapping of a body of text to mathematical unit quantifying the degree of readability. It is the basis of readability assessment. Readability assessment has diverse use cases and applications, such as helping to people with disabilities and also facilitate choosing of learning material for second language learners~\cite{aluisio2010readability}. \par
Learning-based readability assessment refers to those approaches for assessing complexity of a piece of text based by training machine learning (ML) models. The main advantage of the learning-based techniques compared to traditional readability formula (e.g., \textit{Flesch–Kincaid} readability test~\cite{kincaid1975derivation} and \textit{Dale–Chall} readability formula~\cite{mcclure1987readability}) is that they can be trained based on the target group which subjected in the training data. Moreover, to make the the problem of measuring readability easier to solve, traditional readability formulas mainly focus on lexical and syntactic features (e.g., word length and sentence length)~\cite{DBLP:journals/coling/MartincPR21}. On the other side, a range of more diverse features are considered in learning-based approaches. \par
In this paper we proposed two models for automatically assessment of readability of German text based on BERT transformer-based pre-trained language model~\cite{DBLP:conf/naacl/DevlinCLT19}. In our first model, we used GBERT~\cite{DBLP:conf/coling/ChanSM20} (German BERT) to extract features from German text that are used as input for another Recurrent Neural Network (RNN) on top. In the second model, we fine-tuned GBERT based on the training data. We compared the obtained results with a baseline model based on Random Forest method~\cite{DBLP:journals/ml/Breiman01} that proposed in \cite{DBLP:conf/qomex/NaderiMK019}. \par 
Our main contributions in this paper can be summarized as follow:

\begin{itemize}
    %\item Compiling and open-sourcing\footnote{http://shorturl.at/dtD58} a test data  that contains 320 sentences from German \textit{Wikipedia} articles that are scored by an average number of 26 annotators in Level A2-B1. 
    \item Compiling a test data  that contains 320 sentences from German \textit{Wikipedia} articles that are scored by an average number of 26 annotators in Level A2-B1. 
    \item Proposing a text readability assessment model for German text based on transfer learning techniques.
\end{itemize}

The rest of the paper is organized as follow: Section~\ref{sec:relatedwork} presents a number of recent research on text readability and complexity assessment and related tasks in NLP. The used dataset for training and testing the models are thoroughly described in Section~\ref{sec:dataset}. The proposed models and the obtained results are explained in Sections~\ref{sec:proposedmodels} and~\ref{sec:results}, respectively. Finally, discussions and ideas for the future works are highlighted in section \ref{sec:conclusion}. \par

\section{Related Work}
\label{sec:relatedwork}
In this section we briefly describe a number of recent research on automated text readability analysis in German, as well as in English text. \par
In a classical machine learning based method, Xia et al. proposed a text readability assessment system for second language learners~\cite{DBLP:conf/bea/XiaKB16}. To train a supervised ML model, they extracted various types of features from input text which includes lexico-semantic features, parse tree syntactic features, language modeling features, and discourse-based features. The obtained results from their experiments show the SVM classifier could achieve an accuracy of \textit{0.80} and Pearson correlation of \textit{0.9} on the \textit{WeeBit} dataset~\cite{vajjala2012improving}. \par
In a deep learning based approach, Martinc et al. tested a set of neural supervised and unsupervised architectures on three benchmark data in English for the task of readability evaluation~\cite{DBLP:journals/coling/MartincPR21}. They also generated a new Slovenian readability corpus from school book contents. As the unsupervised approach, they tried to develop a new readability formula that can outperform traditional readability formulas by relying on neural language model statistics. Regarding the neural supervised model, they used \textit{BiLSTM}, Hierarchical attention networks (HAN), and transfer learning approaches to train the models. Their obtained results show that the proposed unsupervised readability measure is adaptable, robust, and transferable across languages. \par
Mohammadi et al. proposed a deep reinforcement learning model for readability measuring~\cite{DBLP:journals/corr/abs-1912-05957}. They tested the proposed model on three English and Persian readability corpora. They could skip the feature extraction phase that is essential in the classical machine learning based models, by using the GloVe embedding~\cite{DBLP:conf/emnlp/PenningtonSM14} and statistical language models. In order to extract features, they fed the raw text into a Convolutional Neural Network (CNN) and extracted beneficial features from a piece of text. The obtained results show that the model could outperform the other approaches in Persian text readability dataset and also could achieve competitive accuracy on English data (i.e., the \textit{Weebit} dataset~\cite{vajjala2012improving}). \par 
In a recent effort for classifying German text into different categories, based on text difficulty for language learners, Sz\"ugyi et al. developed a classifier based on linguistic features extracted from the texts~\cite{DBLP:conf/konvens/SzugyiEBS19}. They extracted similar feature as~\cite{DBLP:conf/bea/XiaKB16}, and also error measures and N-grams from input text. To train and evaluate the proposed classification model, they combined five different German readability assessment sources. Their results show that morphological features plays the most important role in accurately classifying the input text into the corresponding difficulty level. Moreover, some of the syntactic and lexical features were also given a high weight by the used ML algorithm. \par
Weiss et al. proposed another learning-based model for complexity modeling for text readability across language~\cite{weiss-etal-2021-using}. They trained different ML models on extracted features from German and English texts. They extracted 312 features includes surface length, syntactic complexity, lexical complexity, morphological complexity and so forth. Feature extraction and selection leads to decreasing the number of features to 301 for model training. Among the ML models that have been trained on the data, SVM could outperform the others and achieve the best results. They achieved the F1 scores of \textit{94.0}, \textit{89.6}, and \textit{92.6} for elementary, intermediate, and advanced texts, respectively. \par

To the best of our knowledge, the proposed model in this paper is the first German model based on transferring knowledge from pre-trained language models to the specific domain of readability evaluation. In the both proposed models we skipped the time-consuming and resource-intensive process of feature engineering.

\section{Dataset} 
\label{sec:dataset}
In this section we thoroughly describe the dataset. All the sentences in both training and test dataset have been collected from \textit{Wikipedia} articles. All sentences have been evaluated by German learners in level A2-B1.  

\subsection{Training Data}
To train the models for the experiments, we used \textit{TextComplexityDE} dataset~\cite{DBLP:journals/corr/abs-1904-07733}. \textit{TextComplexityDE} is a text complexity detection dataset that is compiled for the task of German readability assessment. It contains scores for \textbf{Complexity}, \textbf{Understandability}, and \textbf{Lexical Difficulty}, for 1000 German sentences. All the scores are in range of 1 to 7. \par
\begin{figure}
\centerline{\includegraphics[width=0.65\textwidth]{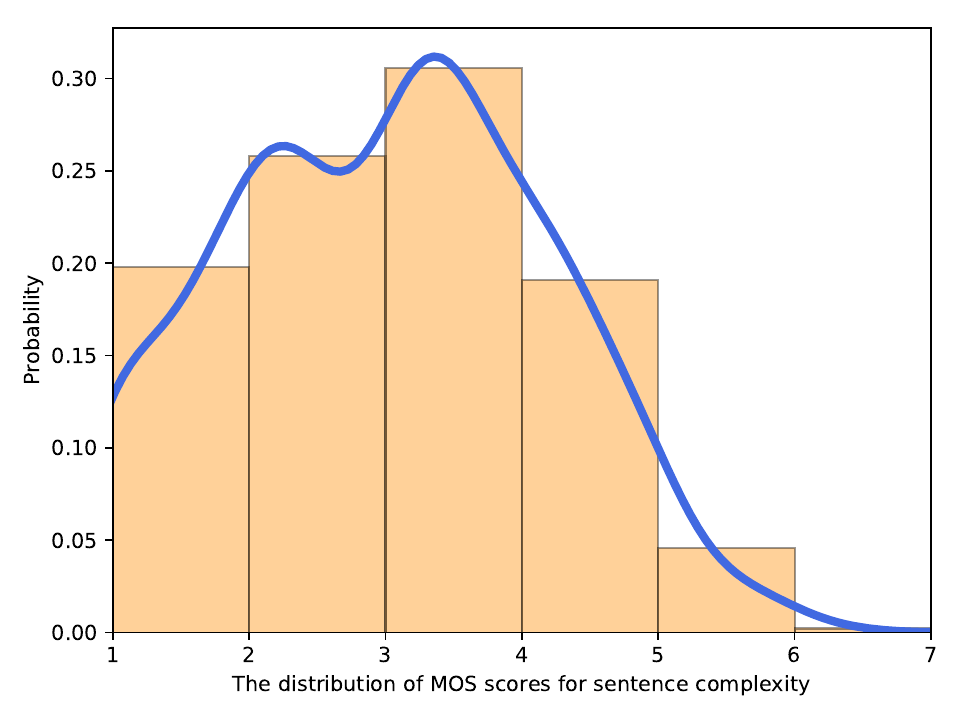}}
\caption{The distribution of MOS values in the training data} 
\label{fig:1}
\end{figure}
\begin{figure}
\centerline{\includegraphics[width=0.65\textwidth]{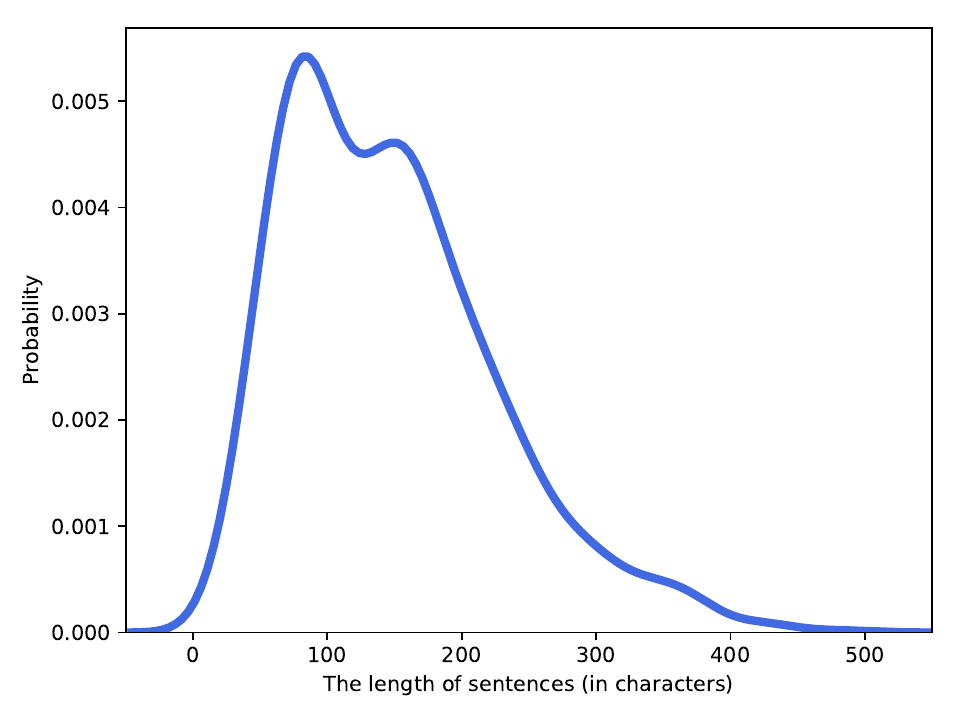}}
\caption{The length distribution of sentences in the training data (in characters)} 
\label{fig:2}
\end{figure}

For this research we used the \textbf{Complexity} scores of 900 sentences to train the models. The Complexity of each sentence in the training data is annotated by a minimum number of 3 and maximum number of 18 annotators. The average number of the opinion scores for each sentence in the data is 10 votes. The distribution of Mean Opinion Score (MOS) for Complexity is presented in Figure~\ref{fig:1}. Also, the distribution of the length of sentences (in character) in the training dataset is depicted in Figure~\ref{fig:2}. As it is highlighted in the figure, most of the sentences are 100 to 200 characters long, while there are a few sentences that are longer than 400 characters. 

Sentences contained in the training dataset were collected from German Wikipedia articles~\footnote{http://de.wikipedia.org} and the Leichte Sprache (Simple language) dataset
developed by Klaper et al.~\cite{DBLP:conf/acl-pitr/KlaperEV13}. More details on the training data is presented in~\cite{DBLP:journals/corr/abs-1904-07733}.

\subsection{Test Data}
The ratings for the test dataset are collected in four different experiments. For each experiment, 100 sentences were complied, 80 from 18 different Wikipedia articles, and 20 sentences were shared between all experiments and taken from the TextComplexityDE dataset. Participants are recruited through online German learner groups in stoical media and also Language schools. For online participants, there was a short mandatory listening and comprehensive language test to make sure they have basic to intermediate knowledge of German language. In addition, online sessions are divided to smaller test sessions in which participant rated 11 sentences in one session. One out of the eleven sentences was a gold standard question (i.e. a question which its answer is obvious and known to the experimenter) which was used for removing submission of participants who are not concentrated. We used a same 7-point Likert Scale as it was used during creation of the TextComplexityDE dataset.

In the data cleansing step, all submission  1) with wrong answer to the gold standard question, 2) which failed in language test, 3) with specific click patterns (i.e. small variance) or those being too fast are removed. For each sentence, Mean Opinion Score (MOS) is calculated by calculating the arithmetic mean over the all ratings provided for that sentence. Using the 20 shared sentences in each experiment, a first-order mapping function for MOS values from each experiment to the MOS values of TextComplexityDE dataset are fitted. It is to remove the well-known bias and gradient between different subjective test. The final test dataset, includes 320 new sentences from 18 Wikipedia articles which rated with minimum 16 participants.

\section{Proposed Models}
\label{sec:proposedmodels}
In this section we present our three proposed models for readability assessment in German text. We start with the baseline model which developed based on classical machine learning techniques, and then we continue with more advanced models based on recently developed pre-trained language models. \par

\subsection{Baseline Model}
As the baseline, we developed a regression model based on Random Forest method~\cite{DBLP:journals/ml/Breiman01}. 
For this purpose, 73 linguistic features have been extracted, grouped in traditional, lexical and morphological features. Feature engineering approaches are employed to select more informative features. \par
After extracting linguistic features, we applied different feature selection techniques such as omitting features with much number of missing values (result in removing 32 features), removing features with high Pearson correlation (result in removing 3 features) and recursive feature elimination (result in removing 18 features). As a result, a total number of 20 features are used to train the Random Forest model to predict readability score of sentences. \par
More details about the baseline model, including the hyper-parameters and the pre-processing steps that have been used in this model are presented in~\cite{DBLP:conf/qomex/NaderiMK019}

\subsection{BERT Model}
In this section we explain our models based on the BERT pre-trained language model~\cite{DBLP:conf/naacl/DevlinCLT19}. Our first model is based on extracting features from the BERT model, in which we do the vectorization of input text by using BERT model. On the other hand, the fine-tuning approach has been used in the second model to slightly change the BERT model's weights based on the data for the experiments. \par 
All the implementations in this section have been done using the \textit{HuggingFace} transformers library~\cite{DBLP:conf/emnlp/WolfDSCDMCRLFDS20} and the \textit{Pytorch} framework~\cite{DBLP:conf/nips/PaszkeGMLBCKLGA19}.

\subsubsection{BERT for Feature Extraction}
In this model we employed BERT to extract features from raw input text. For this purpose, we fed the sentences into the German BERT model (we used \textit{gbert-base}\footnote{https://huggingface.co/deepset/gbert-base}~\cite{DBLP:conf/coling/ChanSM20} model). The weights from the last hidden layer are used as the corresponding vectors of the input tokens. This way, the input tokens from the raw German sentence are converted into vectors based on the context in which a token is represented. \par
\begin{figure}
\centerline{\includegraphics[width=0.55\textwidth]{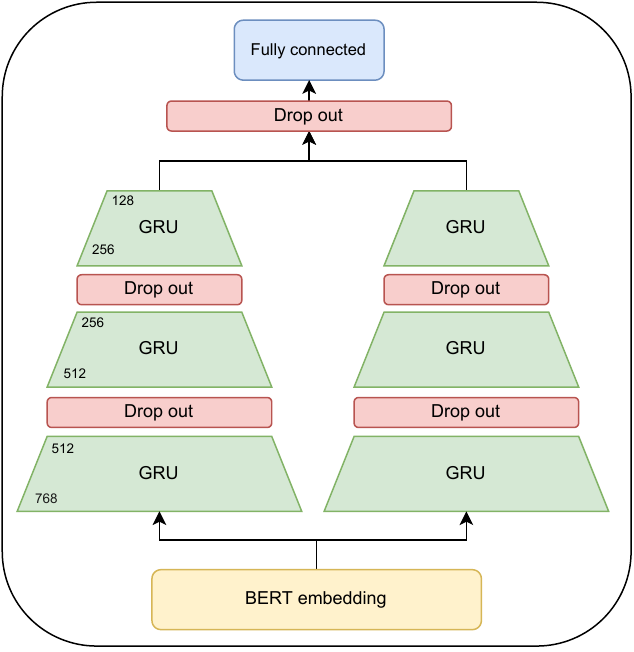}}
\caption{The architecture of the proposed model based on BERT for feature extraction} 
\label{fig:3}
\end{figure}

The resulting embeddings from BERT are fed into multiple Gated Recurrent Units (GRUs)~\cite{DBLP:conf/ssst/ChoMBB14} as depicted in Figure~\ref{fig:3}. To train the model the Adam optimizer~\cite{DBLP:journals/corr/KingmaB14} with learning rate of \textit{1e-3} has been used. Also, we used the batch size of \textit{128} and dropout probability of \textit{0.5} in the experiments.

\subsubsection{Fine-tuning BERT}
The fine-tuning of pre-trained language models has shown promising results in different NLP tasks in recent years~\cite{DBLP:conf/acl-mrqa/SuXWXKLF19,DBLP:conf/icon-nlp/PrabhakarM020}. As a result, for the third model, we developed a BERT based architecture to fine-tune BERT for the readability assessment task. \par
Our fine-tuning based model follows the architecture for sequence classification as it is described in~\cite{DBLP:conf/naacl/DevlinCLT19}. The processes of predicting the MOS value for an input sentence consists of the following steps:

\begin{enumerate}
    \item Tokenization: It includes splitting sentences into tokens and adding special tokens related to the model, and also padding of the sentences.
    \item Passing the token ids through the twelve GBERT layers.
    \item Pooling: It includes the extraction of the last hidden state of the [CLS] token. It passed through a dense linear layer by applying a \textit{Tanh} activation.
    \item Regression: It includes a dropout layer and finally receiving the output float value for the readability score.
\end{enumerate}

\begin{figure}
\centerline{\includegraphics[width=0.45\textwidth]{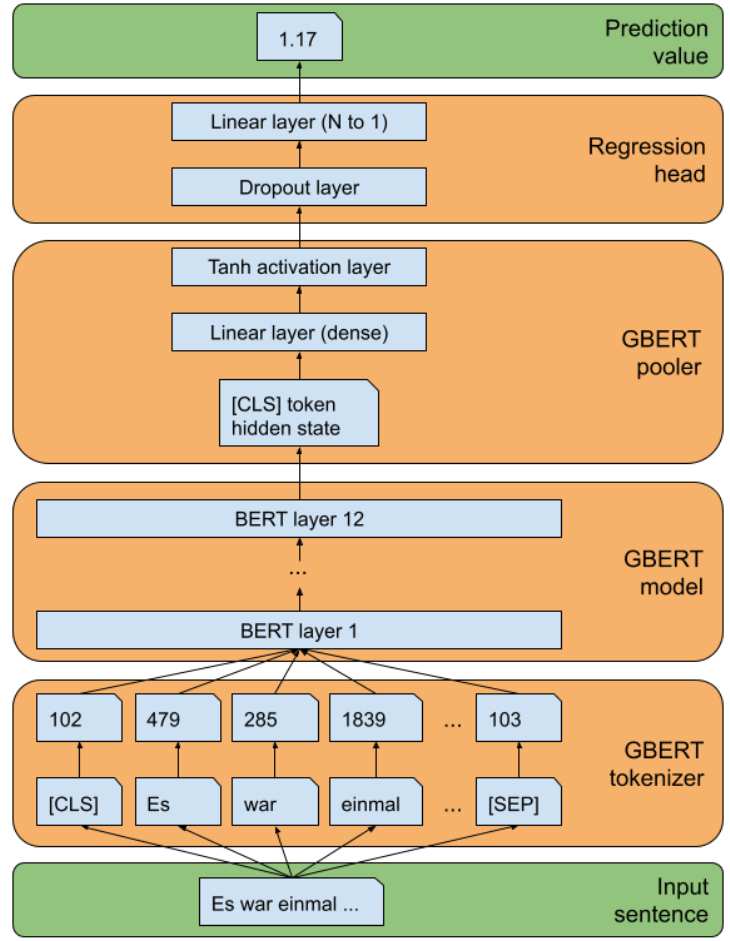}}
\caption{The architecture of the proposed model based on BERT for fine-tuning} 
\label{fig:4}
\end{figure}

We used \textit{gbert-base}~\cite{DBLP:conf/coling/ChanSM20}, for fine-tuning with \textit{AdamW} optimizer~\cite{DBLP:conf/iclr/LoshchilovH19}. Regarding hyper-parameters, the model was fine-tuned in \textit{3} epochs with learning rate of \textit{5e-5} and the batch size of \textit{16}. The architecture of the model is presented in Figure~\ref{fig:4}

\section{Evaluation of Results}
\label{sec:results}

\subsection{Evaluation Metric}
For evaluating the performance of the models, the Root Mean Square Error (RMSE) metric, Pearson and Spearman correlation coefficient are used. 
\subsubsection{RMSE}
shows the root of average squared difference between the estimated values (readability scores) and the actual value, as presented in the following equation. It is a common metric for regression analysis.

% \begin{equation}
% RMSE = \sqrt{\frac { \sum_{i=1}^N {(y_i - \widehat y_i)}^2 }{N}}
% \end{equation}
% where $y_i$ is ith actual value, $\widehat y_i$ is the ith predicted value and $N$ is the number of data points.

\subsubsection{Pearson correlation}
measures the linear correlation between the predicted values and the actual ones for the sentences in the test set.

% \begin{equation}
% r = \frac { \sum_{} {(y_i - \overline{y_i} )(\widehat y_i - \overline{\widehat y_i} )}}{\sqrt{\sum_{} {(y_i - \overline{y_i} ) )^2} \sum_{} {(\widehat y_i - \overline{\widehat y_i})^2}}}
% \end{equation}
% where $r$ is Pearson correlation coefficient, $y_i$ is ith actual value, $\overline{y_i}$ is the mean of actual values, $\widehat y_i$ is the ith predicted value, and $\overline{\widehat y_i}$ is the mean of predicted values.

\subsubsection{Spearman rank correlation}
is correlation coefficient based on ranking the data and then calculating the correlation on the ranked data.

\subsection{Results}

\begin{figure*}[pt] 
   \centering
   \begin{subfigure}[b]{0.3\textwidth}
     \centering
     \includegraphics[width=\textwidth]{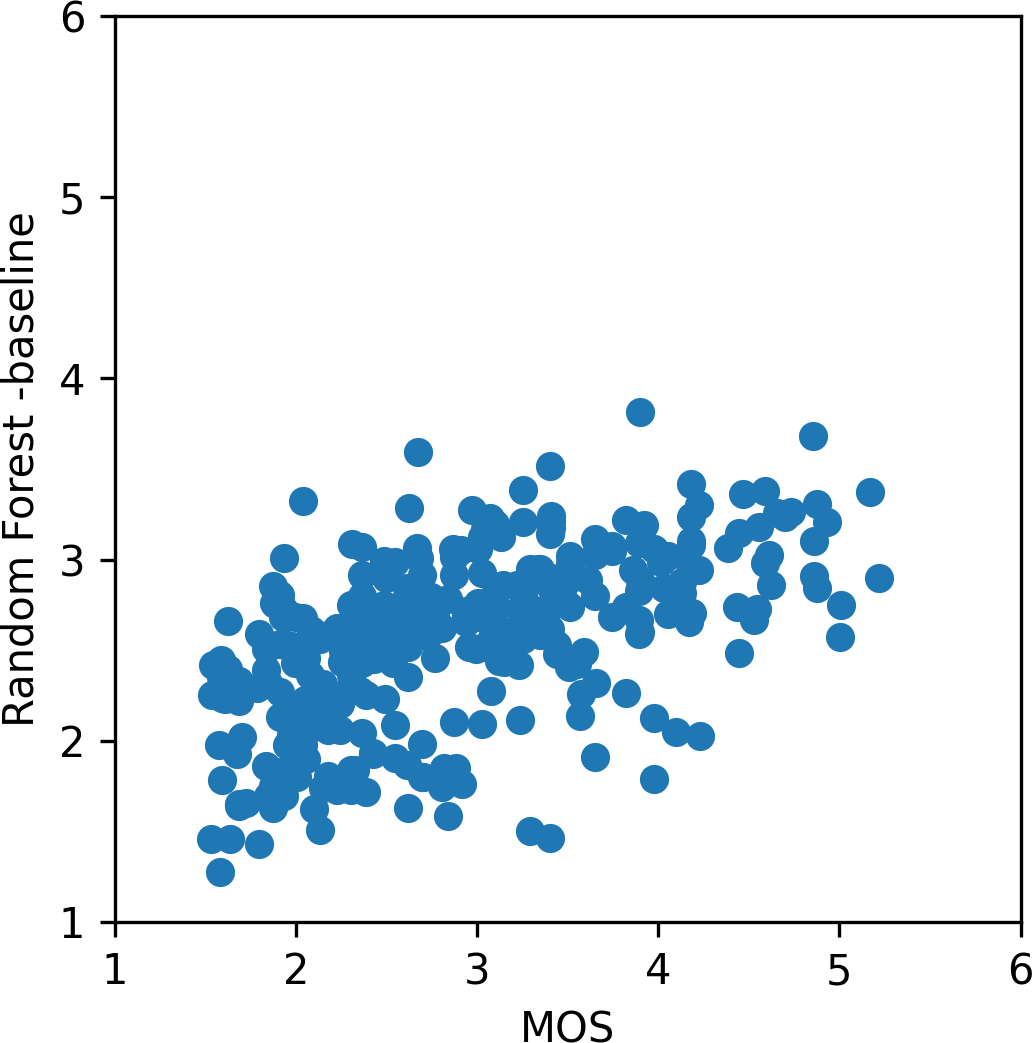} 
     \caption{} 
     \label{figcm:a} 
   \end{subfigure} 
   ~
   \begin{subfigure}[b]{0.3\textwidth}
     \centering
     \includegraphics[width=\textwidth]{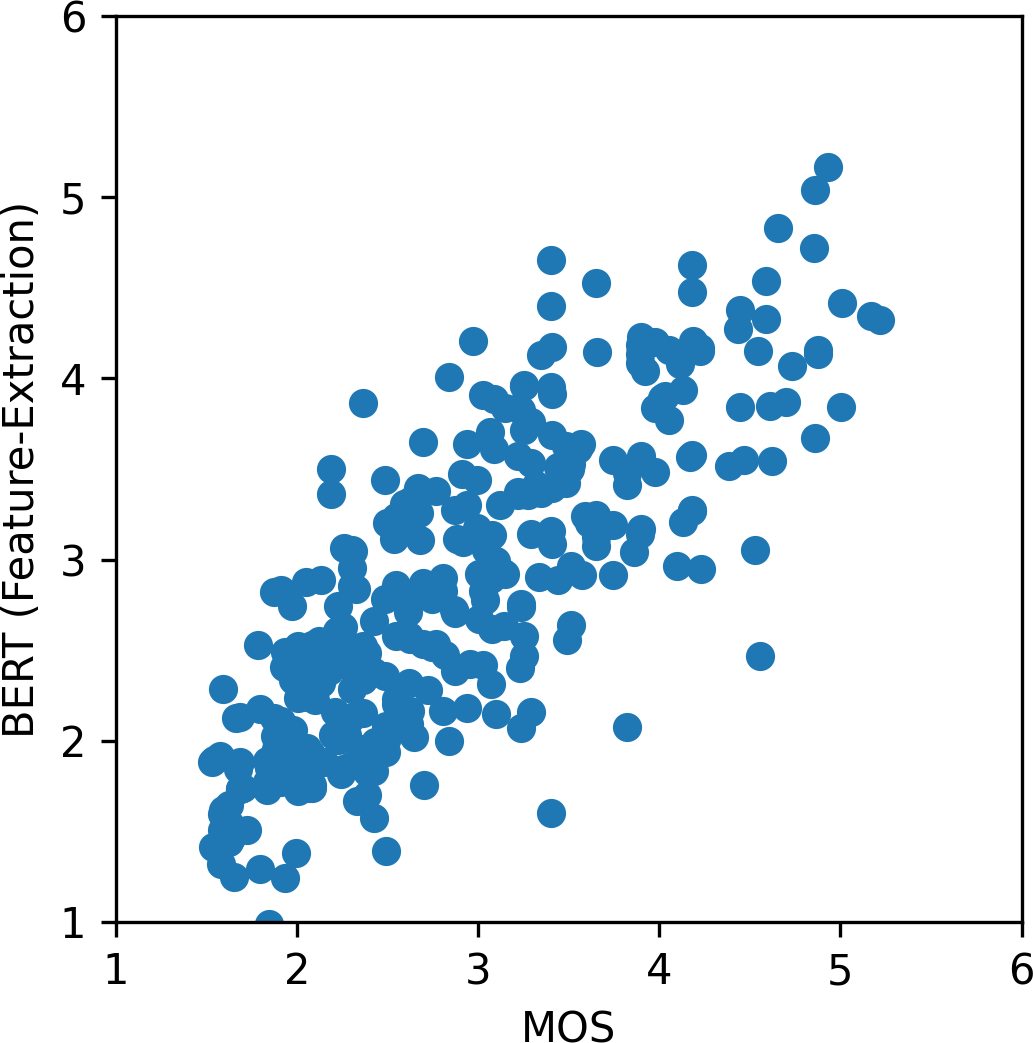}
     \caption{}
 	\label{figcm:b} 
   \end{subfigure} 
   ~
  \begin{subfigure}[b]{0.3\textwidth}
       \centering
     \includegraphics[width=\textwidth]{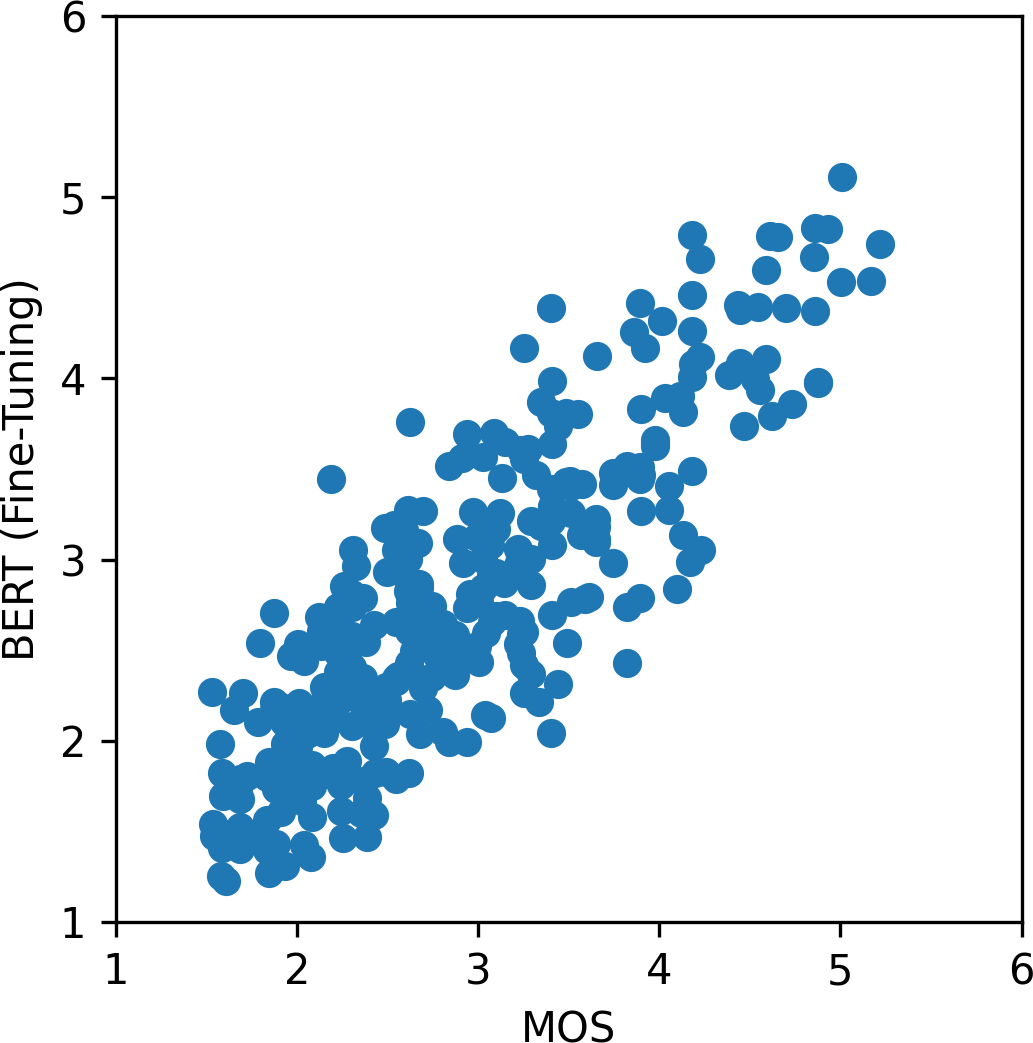}
     \caption{}
 	\label{figcm:c} 
   \end{subfigure} 
   
   \caption{Distribution of predicted MOS values by \textbf{a)} Random Forrest (Baseline), \textbf{b)} BERT (Feature Extraction), and \textbf{c)} BERT (Fine-Tuning)}
   %different models and the subjective test.
   \label{fig:result:mos} 
 \end{figure*}

The obtained results by the three models on the test data are presented in Table~\ref{tab:1}. As highlighted in the table, fine-tuning of the BERT model outperforms the baseline and the feature extraction based model, in \textit{RMSE} and \textit{Pearson correlation} metrics. Distribution of ratings and predicted values from different models are illustrated in Figure~\ref{fig:result:mos}.
\par

\begin{table}
\caption{The obtained results (\textit{RMSE}, \textit{Pearson} and  \textit{Spearman} correlation coefficient)}
\begin{center}
\label{tab:1}
\begin{tabular}{|l c c c|}
\hline
\textbf{Model} &  \textbf{RMSE} & \textbf{Pearson} & \textbf{Spearman}\\
 &   & \textbf{correlation} & \textbf{correlation}\\
\hline
Random Forest (baseline) & 0.813 & 0.551 & 0.581\\
BERT (Feature-Extraction) & 0.557 & 0.796 & 0.799\\
BERT (Fine-Tuning) & \textbf{0.483} & \textbf{0.86} & \textbf{0.844} \\
\hline
\end{tabular}
\end{center}
\end{table}

On the other side, although the feature extraction model achieved promising results, it is far less accurate than the fine-tuning approach in predicting the readability of sentences. Finally, our baseline model achieved approximately the same performance on the test data as the reported results in~\cite{DBLP:conf/qomex/NaderiMK019} from the same model.

\section{Conclusion and Future Works}
\label{sec:conclusion}
In this paper we proposed a transfer learning-based model for text readability assessment in German text. We tested classical machine learning models based on extracting features from text,  and the transformer-based pre-trained language models (e.g., BERT) on two separated test datasets. Our findings show that fine-tuning the BERT model can outperform the other approaches. \par
As future work, different language models (e.g., XLNet and GPT-2) can be tested on the dataset to better validate our findings in this paper. Moreover, since the dataset includes Understandability and Lexical Difficulty scores in addition to the Complexity, a multi-task learning experiment can be applied on the data to measure the impact of the other scores on improving the overall performance of the readability assessment.

%\section*{Acknowledgment}
%%%%%%%%%%%
%To thanks Advanced Project students for compiling the test set. \par
%We should un-comment this section for the camera ready version.
%%%%%%%%%%
%We thanks students of the Advanced Projects course at Technische Universit\"at Berlin (Quality and Usability lab) in summer semester of 2021 to compile a part of the test data as one of the tasks in their projects. \par 

\bibliographystyle{splncs04}
\bibliography{textcomplexity}
\end{document}